# Lessons from a human-in-the-loop machine learning approach for identifying vacant, abandoned, and deteriorated properties in Savannah, Georgia

Xiaofan Liang[1*], Brian Brainerd[2], Tara Hicks[3], Clio Andris[4]

[1]University of Michigan – Ann Arbor, Ann Arbor, United States
[2]City of Savannah, Savannah, United States
[3]City of Savannah, Savannah, United States
[4]Georgia Institute of Technology, Atlanta, United States
[*]Corresponding Author, 2000 Bonisteel Blvd, Ann Arbor, MI, 48109, xfliang@umich.edu

## Abstract

Addressing strategies for managing vacant, abandoned, and deteriorated (VAD) properties is important for maintaining healthy communities. Yet, the process of identifying these properties can be difficult. Here, we create a human-in-the-loop machine learning (HITLML) model called *VAD*ecide and apply it to a parcel-level case study in Savannah, Georgia. The results show a higher prediction accuracy than was achieved when using a machine learning model without human input in the training. The HITLML approach also reveals differences between machine vs. human-generated results. Our findings contribute to knowledge about the advantages and challenges of HITLML in urban planning.

## Introduction

Over time, neglected structures and land in communities degrade into unusable or unsafe infrastructure. This process adversely affects public health and welfare and can increase crime and lower property values. To protect and support growth in communities, local jurisdictions try to identify these vacant, abandoned, and deteriorated (VAD) properties. However, identifying VAD properties, i.e., recording their location, is not a simple task: there is no national database nor a standardized definition with which to detect VAD properties, and as such, cities typically rely on block-by-block field surveys to count VAD properties (Mallach 2018). Often, housing planners drive around the neighborhoods to visually identify and record these properties, or they rely on calls from neighbors to the civic hotline to report problems with properties in their communities[1]. Some cities have used techniques such as spatial decision support systems (SDSSs) and machine learning (ML) models to find VAD properties (Hill et al. 2003; Appel et al. 2014; Hillenbrand 2016; Reyes et al. 2016) but reports of these efforts emphasize the technology itself rather than the combination of human, technology, and data. Minimizing or ignoring the human aspect of these decision-making statistical models is dangerous because the models can provide suggestions that could counter a practitioner's commonsense or what is known to be helpful for a community.

---

[1] Brian Brainerd, Discussion with authors, April, 2021.

Accordingly, this research has two objectives. The first objective is to design a workflow that integrates human expertise into a machine learning model to efficiently identify VAD properties from a dataset of parcels with parcel attributes. The outcome of this objective is a human-in-the-loop machine learning (HITLML) model that we call *VADecide*. The second objective is to measure how *VADecide* improves or does not improve upon (a) a simple, classic machine learning (ML) model using existing datasets (i.e., the "ML model" in this paper) and (b) the city's current workflow based on their expert knowledge. An improved model will be more *efficient* (requiring fewer training samples and giving more predictions), more *accurate* (yielding a higher prediction accuracy rate), more *reliable* (recalling a similar set of properties that were previously discovered using field surveys and U.S. Post Office (USPS) vacancy records), and more *robust* (capturing all features of VAD properties).

To meet these research objectives, we partnered with the City of Savannah, Georgia, in collaboration with four housing officials from Savannah's Housing and Neighborhood Services Department (HNSD) and the Chatham County / City of Savannah Land Bank Authority (LBA). The city government acquires a few dozen VAD properties per year for redevelopment as affordable housing through tax sales, declaration of public nuisance, or eminent domain. To help identify VAD properties, we built a model that inputs a set of parcels (i.e., properties) and outputs a label for each as 'VAD' or 'not VAD'. In the model, each parcel is associated with attributes, such as code violations, tax, and crime, which were collected using census and municipal data records. To help integrate the 'human factor', four housing officials labeled properties by hand in a spreadsheet. The model was trained on these 300 training samples and returns 1,309 predicted VAD properties. We found that *VADecide* is more efficient and scalable than the classic ML model and the city's current workflow. Compared with the ML model, the improved model has a similar model prediction accuracy yet is more reliable (i.e., it has higher agreement rates) when validated with VADs identified through field surveys and USPS vacancy data. The ML model yields a high percentage of VADs with code violations, as it is trained on VADs identified by code violations alone, whereas the properties classified as VADs in VA*Decide* have a wider range of "telling" variables (e.g., tax, code, crime, etc.). Compared with the city's workflow, the HITLML approach identifies more VAD candidates distributed across more neighborhoods, which helps reduce confirmation bias from city planners. The model also associated VAD properties with higher tax delinquency and crime, whereas the city's workflow tends to associate VADs with code violations, which are often manifested as visual features that housing workers see when they visit the neighborhoods.

As such, the contribution of this work is the use of a HITLML model to identify more VAD candidates in diverse neighborhoods, which can help experts create plans for identified parcels and preempt deterioration that adversely affects neighborhoods and residents. The method engages planning experts, and its development unearthed subjectivity in the labeling process (for instance, we found that in some cases, two experts labeled the same property differently). By examining the model's decision process a-posteriori, the experts learned more about their process. While *VAD*ecide's results may not be generalizable to all communities, other planners and researchers may benefit from the lessons learned in this case study, and from the academic-practitioner collaboration that integrates technology with professional knowledge.

Our paper proceeds as follows: We review literature on the topics of vacant, abandoned and deteriorated properties and spatial technologies. We next describe the Savannah, Georgia case study and details on training *VAD*ecide. We then show the results of *VAD*ecide, compared to the ML model and the city's current workflow, and discuss how HITLML aids the city's planning decision-making process.

# Literature Review

### Blighted and VAD Properties

Certain properties in our communities often exhibit problems that give them the distinction as "blighted" properties. Such issues include tax delinquency, prolonged vacancy, structural defects, features that are health hazards (e.g., asbestos, lead paint, or exposed wood shards), overgrowth in vegetation, missing utility infrastructure (e.g., pipes or ventilation systems), and may be subject to flooding (see Luce 2000). The term blight is often used as a legal basis for cities to acquire properties through condemnation (e.g., eminent domain) (Gold, Martin, and Sagalyn 2010). However, this term has been criticized as a stigmatizing term for properties and their neighborhoods (Mallach 2018) and therefore, we use the distinction of Vacant, Abandoned, and Deteriorated (VAD) properties to refer to those that exhibit physical deterioration and neglect (Brachman, 2005; Center for Community Progress 2023).

To our knowledge, there is no known universal definition of VAD (or blighted) properties, and definitions may be tailored to a specific need, such as raising awareness for substandard low-income housing, revitalizing downtown through eminent domain, or revealing the impacts of mortgage crises (Schilling and Pinzon 2016). VAD properties may have lower property values compared with those in the same block, as found in research comparing the sales price of foreclosure properties and the neighboring occupied units (Sumell 2009; Whitaker, Stephan, and Fitzpatrick 2013). Yet in practice, property value can be slow to update and vary significantly based on property attributes and thus not a stable indicator for VADs. Therefore, the definition of VAD must be examined in a local, operational context (see our definition in the Case Study section) and the interactions between VAD characteristics should be considered when distinguishing VAD properties from other properties.

### Causes and Impacts of VAD Properties

A property can deteriorate due to macroeconomic and demographic shifts, a housing market failure, or neglect. In the U.S., many legacy cities (i.e., post-industrial shrinking cities) with sustained job and population loss over the past decade have struggled to support neighborhoods with concentrated vacancy (Mallach, Alan, and Brachman 2013). In other cities, the impact of redlining, a race-based exclusionary housing policy from the mid-20th century, continues to divert investment from predominantly African American neighborhoods and reinforce residential segregation (Rothstein 2017). The housing bubble burst in 2006 and 2007 also led to large-scale foreclosure on properties as the cost to maintain them (e.g., mortgages) exceeded their values (Mallach 2018). At the local level, VAD properties with unclear ownership status or delinquent taxes exceeding the value of the property can be characterized as zombie properties (Alexander 2015, 32), because they have high transaction costs and low property values[2].

VAD properties can disproportionally impact marginalized communities' local housing markets by decreasing neighboring property values and impacting neighbors' quality of life (Whitaker, Stephan, and Fitzpatrick 2013; Mallach 2018). Neighborhoods with many VAD properties are associated with poor school quality (Sun et al. 2019), high crime rates (Branas et al. 2012), higher male unemployment rate (Appel et al. 2014), and slower growth in property sales price (Gilreath 2013). These neighborhoods are also more likely to be home to low-income and African American households (Silverman et al. 2013; Sun et al. 2019) and suffer from declining home ownership and pessimistic perceptions of neighborhood trajectories (Mallach 2021). Geographically, VAD properties also tend to cluster (Hillier et al. 2003; Weaver, Russell, and Bagchi-Sen 2013; Reyes et al. 2016), reinforcing the concentration of income, race, and housing market inequality in marginalized communities and stunting economic mobility. VAD properties also cost millions of USD in lost tax revenue and unrecoverable costs of managing overgrown grass, litter, and illegal dumping; securing open structures; and demolition (Sumell 2009; Immergluck 2016; Mallach 2018).

***Spatial Decision Support Systems for Identifying and Managing VAD Properties***
Early examples of identifying and managing VAD properties with large, digital, spatial datasets include the use of spatial decision support systems (SDSS). SDSSs combine spatial and aspatial data with analytical models and geovisualization to facilitate decision-making in a spatial context (Hopkins et al. 1985; Armstrong et al. 1986). For example, the Philadelphia Neighborhood Information System integrated housing information, web mapping and logistic regression to identify likely abandoned properties (Hillier et al. 2003). Other SDSSs have detected VAD properties from historical records such as code violations or field surveys. The City of New Orleans developed a decision support scorecard system using logistic regression to recommend that city officials sell (on the market or privately) or demolish a property, based on local experts' scoring of multiple criteria related to the property's condition (Hillenbrand 2016). A similar study in the City of Youngstown, Ohio overlayed property characteristics such as vacancy rates and crime rates to create a combined index where properties that 'scored' highest in these collective factors were prioritized for demolition (Morckel 2016). Others have used machine learning models such as random forest, decision tree, and gradient boosting to detect VAD properties and discover how each factor contributes to the prediction (e.g., City of Syracuse in Appel et al. 2014; City of Cincinnati in Reyes et al. 2016). In these case studies, records were often limited to certain neighborhoods. Similarly, the City of Savannah only has 'labels' for a few dozen properties per year. These small sample sizes can deprecate the predictive quality of machine learning models, which typically require larger training datasets, posing a challenge to adopting machine learning models and SDSSs at the municipal level.

---

[2] Brian Brainerd, Discussion with authors, September, 2020

Most case studies are designed to involve planners before or after building the machine learning models (Hill et al. 2003; Appel et al. 2014; Reyes et al. 2016; Hillbrand 2016 is an exception), rather than consulting planners in tasks such as labeling training data and auditing the model's decision mechanisms. As such, planners may be reluctant to adopt the technology due to a lack of trust in and understanding of the model's innerworkings.

### *Human Engagement in Urban Artificial Intelligence*

Traditional supervised machine learning processes involve training a model on big data that has been classified (e.g., whether a property is labeled as VAD) and iterating through multiple probability and statistical tests to predict classes for new data (e.g., properties with unknown VAD labels). In contrast, people use human-in-the-loop machine learning (HITLML) to create new training data from small samples, audit the decision mechanisms behind the algorithms, and calibrate human practices (Holzinger 2016; Zhou, Jianlong and Chen 2018; Monarch 2021). For instance, human experts can carefully examine the results of uncertainty sampling, which selects samples that are most uncertain for machine judgments, reflect on the causes of uncertainty, and then provide the appropriate labels (human judgements) to help the machine learn. Thus, HITLML is especially beneficial in scenarios where there is no large, reliable, labeled dataset and where the complexity of the subject matter necessitates a mutual learning dynamic between the humans and machines, rather than supervision or automation.

The human-in-the-loop approach is part of a rising urban AI movement that outlines the power of technology to govern and serve as a voice for communities (Boehner, Kirsten, and DiSalvo 2016; Loukissas 2019; D'Ignazio, Catherine, and Klein 2020). Technology can encourage civic participation in projects that range from authoritarian, with little public input, to citizen-controlled, in partnership with the public (Arnstein 1969; Cardullo, Paolo, and Kitchin 2019). Others argue that articulating issues, such as design goals, from different stakeholders' perspectives (Le Dantec 2016) and collaborating with expert teams to build, quantify, ground-truth, share, and visualize data (Williams 2020) facilitates the use of technology for public good. While these prior studies call for human-AI collaboration in urban planning, critical discussions around mechanisms and activities for effectively engaging planners and denizens with AI are still developing (Wilson 2022). There are also arguments in urban planning that advocate for a reorientation of technocratic planning toward communicative and collaborative traditions (Mattern 2021). Specifically, a growing number of studies have operationalized HITLML in urban planning: Zheng, Zhibin and Sieber (2021) applied the technique to topic modeling on the text of a corpus of grant proposals on the topic of smart cities, and Anwar (2022) followed suit for land cover mapping applications. Yet, there do not appear to be examples of the benefits of HITLML as applied to issues of housing.

Our research attempts to fill these gaps by proposing an explorative HITLML approach to elevate human presence in machine learning and evaluate the advantages and challenges of a HITLML approach in a planning context, especially for identifying VAD properties.

## Case Study

The City of Savannah is a historic city (city pop 147,780 (U.S. Census, 2020)), a popular tourist destination, and home to the Savannah Port, one of the busiest seaports in the United States (International Trade Administration, n.d.). Savannah is classified by the Lincoln Institute of Land Policy (n.d.) as a legacy city (i.e., a post-industrial shrinking city) whose population peaked in the 1960s.

The city of Savannah's municipal code defines blighted property as one with two or more of the following conditions, including unhabitable structure, inadequate utilities, safety hazards, environmental contamination, repeated illegal activity, active code violations beyond one year, or crime or public health hazards to properties in immediate proximity (City of Savannah, 2016). While this legal definition helps the city acquire such properties through eminent domain, few acquisitions used strategy and thus these cases do not adequately represent all aspects of VAD conditions that can be used for machine learning.

The city has a growing number of observed residential properties that show signs of deterioration and neglect, as measured through code violations and tax delinquency. In 2019, 1,319 properties had code violations that indicate severe physical deterioration and 1,404 properties had at least three years of tax delinquent history. These properties tend to be in minority-concentrated neighborhoods with a sizable Black population.

In 2019, Savannah allocated $10 million USD to repair and redevelop 1,000 properties into affordable housing over the subsequent 10 years (called the "1K-in-10 initiative") in neighborhoods long neglected or exploited by profit-driven investors (Housing and Neighborhood Service, n.d.). To meet this goal, planning experts from Savannah Housing and Neighborhood Service Department (HNSD) experimented with a data-driven approach in 2018. They acquired spreadsheet data from civic departments, visually examined parcel information on an online platform, normalized and averaged the score of each parcel's records in crime, tax, and code violations, and conducted field surveys to validate vacancy on site. However, this process is time-consuming and labor-intensive. The data collected only capture a snapshot of conditions and the field surveys only cover a few neighborhoods. A transparent, scalable, and sustainable VAD-identification system that also considers local context and interactions between variables could help mitigate these issues.

In Savannah, VAD properties have characteristics that are outlined in Savannah's blight standards (City of Savannah, 2016), and that are likely to ease acquisition efforts (based on HNSD planners' experiential knowledge) in the city's 1K-in-10 initiative. In the past, Savannah acquired only dozens of properties annually and thus lacks a large VAD dataset for traditional machine learning prediction. Therefore, we use human-in-the-loop machine learning workflow to label and audit a small subset of VAD candidates with HNSD planners, which enables VAD prediction on a larger set of properties.

## Data

Independent variables used in the city's current workflow are based on Savannah's legal requirements for government acquisition. Such properties must have either active code violations, crime (Part I and Part II types), and/or tax delinquency to qualify for tax (foreclosure) sales, nuisance abatement, or eminent domain. The two machine learning models include additional independent variables suggested by HNSD staff and theories of VAD characteristics (see specific variable definitions in Table 1 and variable nuances in S.I. Section G).

VAD labels for properties identified in the city's current workflow come from a field survey: HNSD staff drove to a few low-income neighborhoods with many highly ranked VAD properties and rated properties based on visual cues aligned with legal blight criteria. In contrast, the ML model's VAD labels are defined by active code violations, and *VAD*ecide's VAD labels are defined by both data and human expertise (see more details in the next sections). To validate ML model and *VAD*ecide's predictions, we use the city's field survey and vacancy data collected by U.S. Post Service (USPS) (see Table 1 for details).

We acquired parcel-level data for the City of Savannah through our partnership with HNSD staff members. Crime and tax data include records from 2010 to 2019. Code violations data include entries from 2012 to 2019. In the two machine learning models, code violations and crime are defined as the weighted count of incidents: an incident is weighted higher if it is more recent and if the type particularly contributes to VAD (see S.I. Section A for details). Property value was dated to 2019 and USPS vacancy data was acquired in 2021.

**Table 1: Independent, dependent, and validation variables per parcel in the city's current workflow, the simple ML model, and *VAD*ecide**

| Variable | Year | Description | Source | Used in |
|---|---|---|---|---|
| **Independent Variables** | | | | |
| Crime | 2010-2019 | The number of crime incidents (Part I and Part II crime). It is weighted by recency and type in ML model and *VAD*ecide. | Police Department | city's workflow, *VAD*ecide, ML model |
| Drug Crime | 2010-2019 | The number of drug crime incidents weighted by recency. | Police Department | *VAD*ecide, ML model |
| Active Code Violation | 2012-2019 | The number of active code violations. It is weighted by recency and type in *VAD*ecide. | Code Compliance | city's workflow, |
| Delinquent Tax | 2010-2019 | Total amount of delinquent city and county tax and unpaid special assessment. The unit is USD $. Due to collinearity, it is combined with delinquent year as one variable when calculating feature importance. | County Tax Office City Tax Office City Revenue Office | city's workflow, *VAD*ecide, ML model |
| Total Delinquent Years | 2010-2019 | The number of years that the property has tax delinquency or unpaid special assessment. Due to collinearity, it is combined with delinquent amount as one variable when calculating feature importance. | County Tax Office City Tax Office City Revenue Office | *VAD*ecide, ML model |
| Unpaid Special Assessment Tax Pct | 2010-2019 | The percentage of unpaid special assessment in total delinquent tax. | Derived | *VAD*ecide, ML model |
| Property Value | 2019 | Property values estimated by computer assisted mass appraisal. The unit is $1,000. | Savannah GIS Parcel Shapefile | *VAD*ecide, ML model |
| **Dependent Variables** | | | | |
| VADs labeled by field survey | 2019 | Properties identified as VAD candidates in a 2019 field survey by human experts in selected neighborhoods. | Field Survey | city's workflow |
| VADs labeled by active code violations | 2019 | The number of active code violations under designated type "Condemnation", "Vacant Property Clean/Mow", "Unsafe Secure", and "Unsafe Demolition" in code compliance database, which indicates VAD conditions. | Code Compliance | ML model |
| VADs labeled by data + human expertise | 2019 | Properties identified as VAD candidates in the human-in-the-loop workflow by human experts based on independent (input) variables in the model and their experience. | Human experts | *VAD*ecide |
| **Validation/Consensus Variables** | | | | |
| VADs labeled by field survey | 2019 | (repeated from above) Properties identified as VAD candidates in a 2019 field survey by human experts in selected neighborhoods. | Field Survey | *VAD*ecide, ML model |
| USPS vacancy | 2021 | Properties identified as vacant (not collecting their mail) by USPS staff for 90 days or longer. | USPS | city's workflow, *VAD*ecide, ML model |

## Methods

### *A Simple Machine Learning Model*

This paper defines a 'simple' machine learning model (also referred to as the 'ML model') as a supervised model with few or no expert interactions. We labeled VAD properties as those with certain code violations, such as vacant property, need for cleaning/mowing, unsafe demolition, and condemnation. We removed code violation variables from the simple ML's input features because we used code violation to construct the VAD labels. Unlike *VAD*ecide, the ML model learns VAD characteristics from labeled VADs in 2019 and then makes predictions without expert inputs, as is typical in similar studies.

### *Human-in-the-loop Machine Learning (VADecide)*

We created a workflow for *VAD*ecide (Figure 1) to identify/label example VAD properties. Steps 4 through 6 specifically distinguish a human-in-the-loop approach from a simple ML model and from the city's current workflow (see detailed discussion of how the three approaches diverge in each step in S.I. Section B).

## Predicting VAD Properties via Human-in-the-loop Machine Learning Workflow

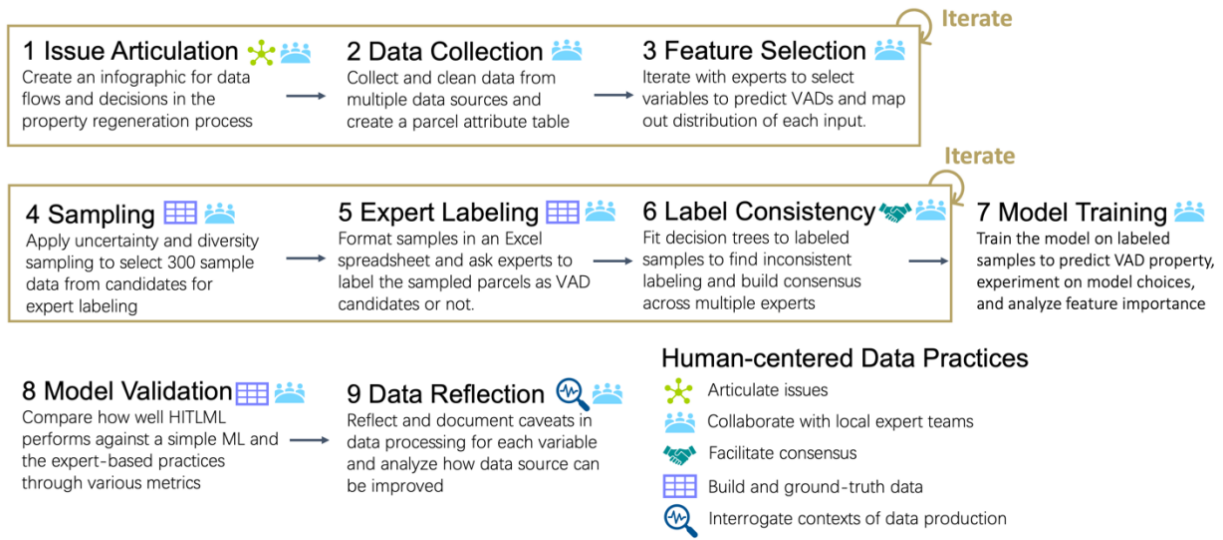

**Figure 1: Human-in-the-loop machine learning (HILML) workflow used to create *VAD*ecide. The symbols represent various human-centered data practices we implemented in each step.**

### *Issue Articulation*

To identify whether and where machine learning interventions help the decision-making process, we first sent a Q&A document (S.I. Section C) to the HNSD experts asking them to document the process of property regeneration from an administrative perspective. They reported that this process involves collecting data, identifying VAD candidates, determining acquisition strategies, acquiring properties, and regenerating the properties for various purposes. We next worked with the team at their offices in Savannah and online to codify how they decide to label a parcel. To help communicate these processes, we collaboratively created an infographic with the team (S.I. Section D); this graphic shows that the biggest challenge is managing decentralized data that flows between institutions and leveraging the data for effective decision-making. By unveiling the decision process behind VAD property identification and acquisition, we realized that machine learning can help identify VAD candidates but is less successful at matching VAD candidates to acquisition strategies (e.g., tax sales, public nuisance, eminent domain), as this process requires contextual knowledge that is difficult to codify into data (e.g., some neighborhoods are more receptive to certain strategies). Thus, we narrowed the goal of the machine learning intervention as VAD identification.

### *Data Collection and Feature Selection*

We created maps of Savannah symbolized by variables listed in Table 1 and examined them in person with the experts to decide on filtering criteria for model input data. This discussion found 5,372 residential structures (single-family, 2-4 family, and townhouse) and lands that have records in drug crime, tax issues, code violations, or fire incidents (required by Savannah city law for government property acquisition) and have little or no flood risk (which was a priority of HNSD). While the expert workflow uses tax, code, and crime data as decision input, the machine learning models take in many variables and report an importance score for each. We experimented with fifteen variables, including building attributes (such as land size, number of bedrooms) and market indicators (such as number of unqualified sales) (see S.I. Section A for all variables), but only kept a minimal list (see Table 1) that combines planners' recommendations and model suggestions for final prediction (Step 2 and 3 in Fig 1).

*VAD Properties Sampling and Labeling*

Since Savannah does not have a database with properties labeled as VAD, we used a mixture of random sampling and active learning techniques (uncertainty and diversity sampling) to create training samples (i.e., which property can be a VAD candidate). Uncertainty sampling selects data points that are close to decision boundaries and thus have the highest uncertainty (Lewis, David, and Catlett 1994), while diversity sampling ensures the geographic and representation diversity in samples (Monarch 2021) (see S.I. Section E for technical details) (Step 4 in Fig 1).

After sampling, we asked a team of four housing experts to classify a subset of properties (n=300) as VAD or not; this serves as a dependent variable for the model's training data (Step 5 in Fig 5). The experts were two males and two females ranging in age from their 30s to 65. They have experience in the field of land and housing for 25 years, 15 years, 12 years and 3 years, and have been working for the City of Savannah for 8 years, 3 years (x2), and 2 years. They had no prior experience in machine learning nor in labeling training data.

We sent each expert a spreadsheet of properties and their characteristics (see S.I. Section E). In the spreadsheet, each column is a parcel, and each row is a parcel attribute. The experts could choose labels (i.e., VAD and Not VAD) from a dropdown menu in the spreadsheet and write comments to clarify how they made the decisions. One expert labeled 150 samples, and the others labeled 50 samples each.

After labeling the training data, we asked experts via e-mail to discuss labels that either conflicted with their professional knowledge or were deemed as highly uncertain by the machine learning algorithm (see S.I. Section E) (Step 6 in Fig 5). We found that properties with very similar conditions were labeled differently, presenting opportunities to highlight implicit assumptions in the labeling process and to streamline the method to ensure equity across the process. To find these properties, we fitted decision trees to labeled data of parcels with and without structures, respectively, and visualized how the model made decisions for each type. For example, if a VAD property encountered multiple splits in the decision tree and was separated from a large group of non-VAD properties, then it was a good candidate for discussion (see S.I. Section D for visualization). When experts labeled a property differently, we presented the decision trees and the labeled samples, and the experts collaboratively agreed upon a resolution. This process helped expose nuances in the interactions of variables, differentiate whether false labels were due to low sample points or human error, and uncover tacit variables being used in the reasoning.

### HITLML Model Training, Interpretation, and Prediction

To create *VADecide*, we used a *random forest algorithm* to classify properties into VAD and not VAD properties (Step 7 in Fig 1). The random forest model captures nonlinear relationships between variables by creating a "forest" of decision trees in which each tree decides on a sequence of variables and variable values to split the data to minimize label differences in each branch (Liaw, Andy, and Wiener 2002). We chose a random forest model because it has higher model accuracy than logistic regression and because VAD status depends on a combination of variables (e.g., crime with tax delinquency); random forest can capture such feature interactions and is 'smart' enough to pick between correlated variables (Basu et al. 2018) (see S.I. Section E).

We used drop-column feature importance and partial dependence plots to experiment with different model settings, including different feature combinations and different numbers of training samples (see S.I. Section E). In the end, our final HITLML model is trained on 300 samples (Land n=105; Structure n=195) with seven features (see Table 1).

### Model Evaluation and Validation

We compared the *VAD*ecide-predicted VADs with ML-predicted VADs and VADs identified by the city's current workflow through five metrics (Step 8 in Fig 1):

Input refers to the number of VADs surveyed by the city or trained in the ML model and VADecide. A higher value means that more data collection is needed to be collected to make inference.

Output refers to the number of VADs found by the city or predicted by the ML model and VADecide. A higher value means that the process returned more VAD candidates.

**Internal accuracy** (ML and *VAD*ecide only) refers to the prediction accuracy trained with 80% of labeled VAD properties and validated with the remaining 20% (reserved from labeled VAD dataset, i.e., test sets). We used cross-validation and out-of-bag scores, two standard metrics in machine learning to measure the accuracy (see S.I. Section F for OOB score results and technical explanations). A higher value indicates better prediction accuracy.

**The external consensus rate** refers to the percentage of properties that are also VAD candidates in other sources (i.e., the field survey and USPS vacancy data); a higher consensus rate indicates more reliable predictions. A higher value means more reliable prediction. Both field survey and USPS data are not perfect representations of the locations of VAD properties; the ideal 'ground-truth' would be VAD properties that were identified and acquired in the past, but this population is too small and not diverse in our case study site.

**Content sensitivity** refers to the percentages of VAD properties dominated by different features. We focused on crime, code violation, tax delinquency, and low property values as VAD types because they are important features for predicting VADs in our random forest model (see S.I. Section F for details on how we determine types). A higher value means that a method is less robust and is likely to retrieve VAD candidates based on a single variable.

# Results

*Comparison between VADecide, Simple ML, and the City's Current Workflow*

Considering input-output, *VAD*ecide is the most efficient among the three, followed by the ML model and then the city's current workflow. *VAD*ecide identified 1,309 VAD properties among 5,372 candidates, using only 300 carefully selected expert-labeled samples. In contrast, the field survey from the city's current workflow contains 890 effective samples and identifies only 192 as VAD candidates.

Next, both the *VAD*ecide and the simple ML model have high internal (prediction) accuracy. The minimum accuracy is *VAD*ecide's prediction on structures (88%) and the maximum is ML's prediction on structures (93%).

Regarding consensus rates, *VAD*ecide exceeds those produced by the ML and the city's typical workflow. For example, *VAD*ecide's predictions overlap with 60.49% VAD land and 63.96% VAD structures reported in the field survey and 66.67% VAD land and 34.61% VAD structures found in the USPS vacancy data. Since neither source represents the ground-truth status of VADs, the accuracy for *VAD*ecide may be even higher. The range of consensus rates for *VAD*ecide (34.61% to 66.67%) is also much higher than the range in the ML model (14.81% to 22.52%), indicating that the human-in-the-loop approach is more reliable, while the ML model performance may be biased by using a single variable (i.e., code violations) to label VAD samples. *VAD*ecide also outperforms the city's current workflow (range 26.15% to 48%).

The three methods also exhibit distinct distributions of predominant VAD features, indicating that they have varying degrees of content sensitivity. This divergence implies that certain VAD characteristics exert greater influence (or bias) than others. For instance, both the *VAD*ecide and the ML model predict more VAD properties with high crime and tax delinquency records, whereas the city's current workflow identifies more properties with low property values. The *VAD*ecide-predicted VADs also have fewer code violations than those identified by the city. These discrepancies may be caused by differences in the human-versus-machine identification process. In the city's *current* workflow, planning experts use low-income neighborhood location and low property value as heuristics for identifying easy-to-acquire VAD candidates. During field surveys, they are also more likely to rely on visual cues, such as dilapidated roofs and broken windows, which tend to be associated with code violations. The simple ML model predicted that VADs have higher rates of code violations than those detected from the city's current workflow and from *VAD*ecide. This reveals that using one VAD feature (in this case, number of code violations) as training properties' VAD labels may cause the ML model predictions to overrepresent properties with this feature. In contrast, *VAD*ecide's results show that the percentage of identified VADs with various features (e.g., crime, code violations, tax delinquency, low properties values) falls between the percentages reported by the ML model and the city's current workflow, indicating that they may be a 'happy medium' between the techniques (Table 2). In comparison to VADs retrieved from the city's workflow, VADecide-predicted VADs are spread more widely across neighborhoods and are distributed more evenly across land and structure parcel types. This outcome may be the result of the city's neighborhood visitation constraints, as field surveys cover a limited set of (low income) neighborhoods, while *VAD*ecide was trained with geographically diverse samples. The wider distribution of *VAD*ecide-predicted VADs mitigates human bias in associating VADs with marginalized communities. VADs predicted by the ML model are also distributed across more neighborhoods than those retrieved when using the city's workflow. Yet, the ML model's predicted properties are concentrated around neighborhoods with available code violation data (see S.I. Section G). For reference, USPS vacancy data have the widest geographic spread, but these data are biased toward vacant structures because structureless properties rarely receive mail.

**Table 2: Comparison of city's current workflow, ML model and *VAD*ecide**

| Metrics | Interpretation | City's workflow | | ML model | | *VAD*ecide | |
|---|---|---|---|---|---|---|---|
| | | Land | Structure | Land | Structure | Land | Structure |
| Input | Total VADs surveyed in the city's current workflow or used to train the ML model or *VAD*ecide. Higher values indicate that more records are needed to make an inference. | 483 | 407 | 183 | 353 | 105 | 195 |
| Output | Total VADs identified in the city's current workflow or that were predicted by ML model or *VAD*ecide. Higher values means that more VADs are identified. | 81 | 111 | 96 | 281 | 472 | 837 |
| Internal accuracy | Percent of ML-predicted VADs (trained with 80% labeled VADs) that are also labeled as VAD in the test set (20% of labeled VADs not used for ML training). Higher values mean better prediction accuracy. | NA | NA | 90.72% | 92.95% | 93.33% | 87.69% |
| External consensus | Percent of VADs identified/predicted that are also found in the city's <u>field surveys</u> (CV score) | NA | NA | 14.81% | 22.52% | 60.49% | 63.96% |
| | Percent of VAD candidates identified/predicted that are also found in the <u>USPS vacancy dataset</u> (CV score) Higher value in either indicates a more reliable prediction output. | 48% | 26.15% | 16.67% | 17.67% | 66.67% | 34.61% |
| Content sensitivity | Percent of identified VADs with <u>crime</u> | 7.4% | 22.3% | 9.4% | 30.6% | 8.3% | 46.5% |
| | Percent of identified VADs with <u>code violations</u> | 34.6% | 42.7% | 82.3% | 63% | 29% | 30.5% |
| | Percent of identified VADs with <u>tax delinquency</u> | 75.3% | 83.5% | 100% | 86.5% | 99.8% | 98.2% |
| | Percent of identified VADs with <u>low property value</u> Higher values means that a higher percentage of VADs have these features and thus a particular method is less robust and is likely to depend on one data source versus others | 96.3% | 83.5% | 86.5% | 71.9% | 90.3% | 68.2% |

**Notes:** CV refers to cross-validation accuracy (see S.I. section F for further explanation). USPS refers to vacancy records compiled by USPS staff. NA values are present because the data collected using the city's current workflow (i.e., field surveys) could only be compared with USPS data, and do not run through the ML learning process.

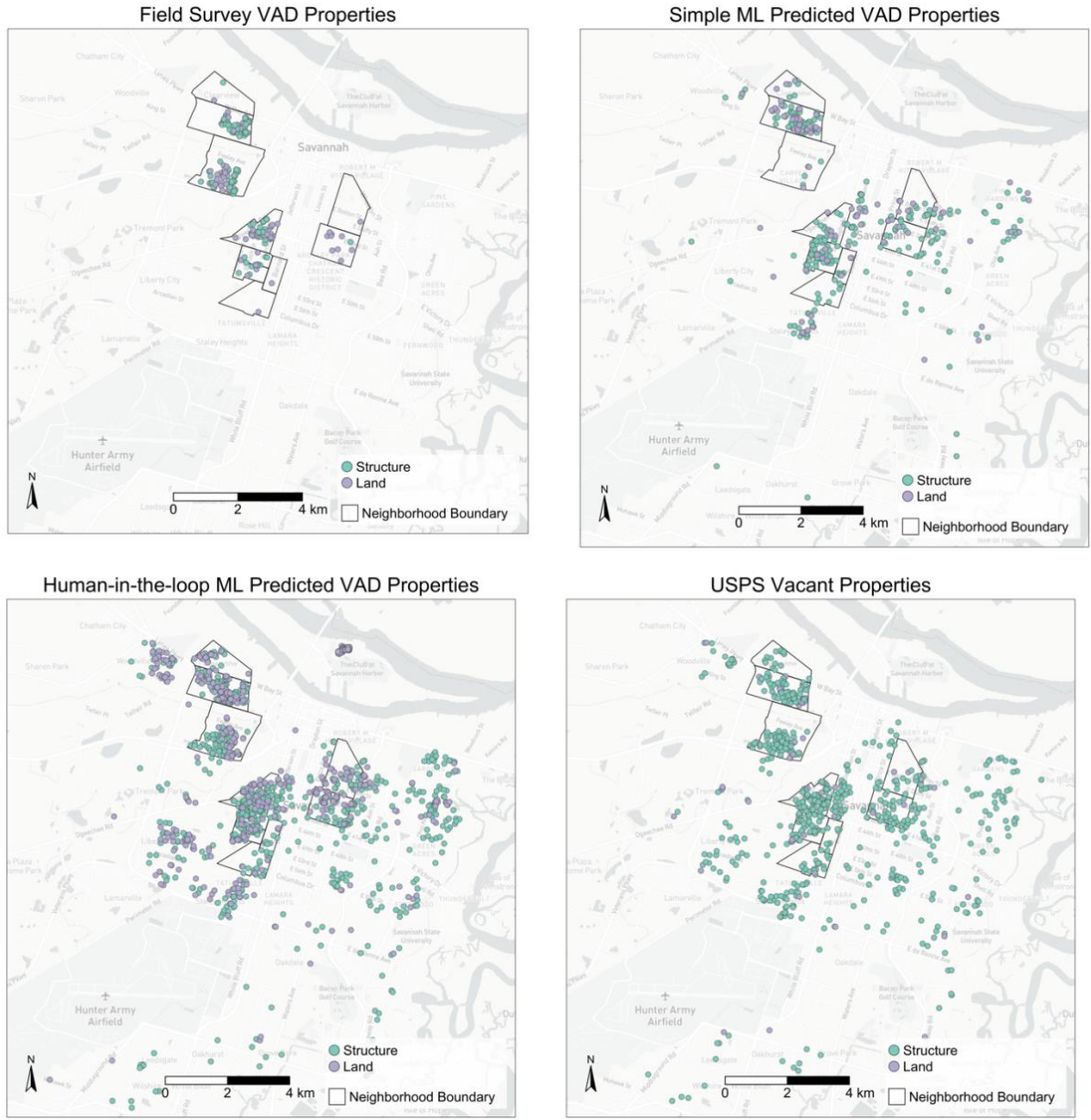

**Figure 2: Geographic distribution of identified/predicted VADs in city's current workflow, ML model, and *VAD*ecide. USPS vacant properties are mapped as an external validation.**

*Reflection*

We met with HNSD staff and shared a table that outlined limitations and notes for each feature in the data collection and processing steps (see S.I. Section G) (Step 9 in Fig 1). Together, we reflected on possible issues with interpretation, generalizability, and suggestions for improving data source quality. For example, in this post hoc examination, we noticed a lack of code cases from 311 calls (a civic hotline in Savannah and a key source of code violation data for the city) in certain neighborhoods. Using linear regression, we subsequently found that neighborhoods with lower household income and higher percentage of African American population correlate with few 311 calls. We subsequently discussed possible ways to encourage more 311 calls from these neighborhods (see S.I. Section G).

## Discussion and Conclusion

This paper presented a human-in-the-loop machine learning approach (*VAD*ecide) that predicts vacant, abandoned, and deteriorated (VAD) properties in Savannah, Georgia. We found that *VAD*ecide-predicted VADs not only have comparable accuracy to the classic ML model but are also more reliable than outcomes from both the ML model and the city's current workflow. We also discovered differences in what *VAD*ecide and HNSD (human experts) would classify as VAD: *VAD*ecide identifies more properties that are tax delinquent, prone to crime, and in neighborhoods with various income levels, while human experts often rely on visual cues from code violations and low property value data. *VAD*ecide's predictions also covered a wider range of neighborhoods beyond the scope of human experts' field survey.

However, our study also has some limitations. First, our model focuses on challenges from the institutional perspective rather than the challenges of the property owners. As such, we did not incorporate the voices of community members, renters, and property owners, which reinforces a transactive, normative mode of governance (where denizens are consumers of public services) rather than a relational mode (where denizens are co-creators). Despite the benefits of integrating expert knowledge into the workflow, the human-in-the-loop approach has been criticized as a method that simply uses people as labor to generate and calibrate data (Janowicz et al., 2022). Second, involving housing experts to contextualize the selection of variables, build training data, adapt the model, and validate the model outcomes is costly. Model accuracy also requires constant data updates, which is a common struggle in many smart city initiatives (Kitchin 2014). Lastly, our model only identifies candidates based on VAD characteristics, but does not suggest how the government should proceed in managing the property. Acquisition in Savannah, and many other cities, is subject to political (e.g., minority communities' concerns with eminent domain) and financial factors (e.g., case-by-case acquisition costs and benefits)[3]. The extra properties identified by *VAD*ecide present opportunities for early rehabilitation and intervention before these properties become uninhabitable. Future studies can investigate whether ML-identified VAD candidates are better for various acquisition strategies (e.g., tax sale, nuisance abatement, eminent domain, arms-length transaction) or regeneration options (e.g., as affordable housing, green space, rehabilitation), as compared with sites identified through fieldwork (see Figure S1 in Appendix for the city's property regeneration workflow).

We recommend the human-in-the-loop model for municipal governments that lack an unbiased, large-scale VAD dataset for machine training, are ready to maintain and upgrade their digital infrastructures, and seek an alternative approach to evaluate their existing workflow. For cities without a consistent definition and acquisition of VAD properties, the HITLML model is beneficial as it requires only a small training sample to generate standardized, accurate, reliable, and robust predictions. The intentional process for human experts to define, deliberate, and dialogue with the HITLML model can also be enlightening and trust-building, even if it is a one-off practice. For instance, Savannah planners found the sample labeling process helpful in organizing their thoughts, making decisions, and reflecting on why they chose certain outcomes for the parcels. Two experts expressed increased confidence and knowledge about deploying the model in future operations. One expert also mentioned that interpreting the differences between human and machine-predicted outcomes made them rethink hidden assumptions in their current workflow.

---

[3] Brian Brainerd, Discussion with authors, April, 2021

The main barriers to consistently deploying the HITLML model in practice include technical and labor costs in maintenance. Critical digital infrastructures may include a central database of all VAD-relevant variables, often collected by separate civic departments, cloud resources to deploy the model, a visualization platform to overlay parcel details and predictions on the map, and an interface that allows humans to validate the predictions. Local governments may also need to consider the labor cost of hosting training workshops for staff to understand, recalibrate, and use the model and its outcomes, and develop guidelines for auditing and communicating results with community members. These challenges have also prevented the City of Savannah from adopting our HITLML model for the long term.

In conclusion, this research describes a human-in-the-loop machine learning approach for classifying land and residential structures as potentially in need of attention. By using a collaborative technique that engages experts while using large datasets and statistical analysis, we generated new insights into how to potentially automate or improve the local government's ability to identify vacant, abandoned, and deteriorated properties. The result is a more reliable method for managing assets in a municipal setting. We suggest that more researchers and practitioners collaborate and incorporate human input and expertise as they develop and test their machine learning applications.

## Acknowledgement

This work was funded by the Georgia Partnership for Inclusive Innovation (PIN).